\documentclass[sn-mathphys,Numbered]{sn-jnl}

\usepackage{graphicx}%
\usepackage{multirow}%
\usepackage{amsmath,amssymb,amsfonts}%
\usepackage{amsthm}%
\usepackage{mathrsfs}%
\usepackage[title]{appendix}%
\usepackage{xcolor}%
\usepackage{textcomp}%
\usepackage{manyfoot}%
\usepackage{booktabs}%
\usepackage{algorithm}%
\usepackage{algorithmicx}%
\usepackage{algpseudocode}%
\usepackage{listings}%
\usepackage{mathtools}
\usepackage{tcolorbox}
\tcbuselibrary{breakable}
\usepackage{hyperref}

\usepackage[backend=biber, maxcitenames=2, sorting=nty]{biblatex}
\usepackage{acro}
\DeclareAcronym{ABIDE}{
  short=ABIDE,
  long=Autism Brain Imaging Data Exchange,
  tag = abbrev
}
\DeclareAcronym{KL divergence}{
  short=KL divergence,
  long=Kullback--Leibler divergence,
  tag = abbrev
}
\DeclareAcronym{kNN}{
  short=$k$NN,
  long=$k$-Nearest Neighbors,
  tag = abbrev
}
\DeclareAcronym{DFT}{
  short=DFT,
  long=Discrete Fourier Transform,
  tag = abbrev
}
\DeclareAcronym{FFT}{
  short=FFT,
  long=Fast Fourier Transform,
  tag = abbrev
}
\DeclareAcronym{KDE}{
  short=KDE,
  long=Kernel Density Estimation,
  tag = abbrev
}
\DeclareAcronym{FFTKDE}{
  short=FFTKDE,
  long=Fast Fourier Transform-based Kernel Density Estimation,
  tag = abbrev
}
\DeclareAcronym{AG}{
  short=AG,
  long=Accelerated Gradient,
  tag = abbrev
}
\DeclareAcronym{CG}{
  short=CG,
  long=Conjugate Gradient,
  tag = abbrev
}
\DeclareAcronym{GMRES}{
  short=GMRES,
  long=Generalized Minimal Residuals,
  tag = abbrev
}

\DeclareAcronym{SCAD}{
  short=SCAD,
  long=Smoothly Clipped Absolute Deviation,
  tag = abbrev
}
\DeclareAcronym{MCP}{
  short=MCP,
  long=Minimax Concave Penalty,
  tag = abbrev
}

\DeclareAcronym{PCA}{
  short=PCA,
  long=Principal Components Analysis,
  tag = abbrev
}

\DeclareAcronym{ICA}{
  short=ICA,
  long=Independent Component Analysis,
  tag = abbrev
}

\DeclareAcronym{GWAS}{
  short=GWAS,
  long=Genome-Wide Association Studies,
  tag = abbrev
}

\DeclareAcronym{GMM}{
  short=GMM,
  long=Generalized Method of Moments,
  tag = abbrev
}

\DeclareAcronym{MRI}{
  short=MRI,
  long=Magnetic Resonance Images,
  tag = abbrev
}
\DeclareAcronym{CRLB}{
  short=CRLB,
  long=Cramer--Rao Lower Bound,
  tag = abbrev
}
\DeclareAcronym{NYSE}{
  short=NYSE,
  long=New York Stock Exchange,
  tag = abbrev
}
\DeclareAcronym{NASDAQ}{
  short=NSADAQ,
  long=National Association of Securities Dealers Automated Quotations,
  tag = abbrev
}
\DeclareAcronym{BSM}{
  short=BSM,
  long=Black--Scholes--Merton,
  tag = abbrev
}
\DeclareAcronym{LASSO}{
  short=LASSO,
  long=Least Absolute Shrinkage and Selection Operator,
  tag = abbrev
}
\DeclareAcronym{SNR}{
  short=SNR,
  long=Signal-to-Noise Ratio,
  tag = abbrev
}

\theoremstyle{thmstyleone}%
\theoremstyle{thmstyletwo}%

\theoremstyle{thmstylethree}%

\begin{document}

\title[{\tt fastHDMI}: Fast Mutual Information Estimation for High-Dimensional Data]{{\tt fastHDMI}: Fast Mutual Information Estimation for High--Dimensional Data
}


\author*[1]{\fnm{Kai} \sur{Yang}}\email{kai.yang2@mail.mcgill.ca}

\author[1]{\fnm{Masoud} \sur{Asgharian}}\email{masoud.asgharian2@mcgill.ca}

\author[1, 3]{\fnm{Nikhil} \sur{Bhagwat}}\email{nikhil.bhagwat@mcgill.ca}

\author[1]{\fnm{Jean-Baptiste} \sur{Poline}}\email{jean-baptiste.poline@mcgill.ca}

\author[1, 2]{\fnm{Celia} \sur{Greenwood}}\email{celia.greenwood@mcgill.ca}

\affil*[1]{\orgname{McGill University}, \city{Montreal}, \state{Quebec}, \country{Canada}}

\affil*[2]{\orgdiv{Lady Davis Institute for Medical Research}, \orgname{Jewish General Hospital}, \city{Montreal}, \state{Quebec}, \country{Canada}}

\affil*[3]{\orgdiv{Origami Lab, Montreal Neurological Institute}, \orgname{McGill University}, \city{Montreal}, \state{Quebec}, \country{Canada}}



\abstract{In this paper, we introduce {\tt fastHDMI}, a Python package for the efficient execution of variable screening for high--dimensional datasets, including neuroimaging datasets. This study marks the inaugural application of three distinct mutual information estimation methodologies for variable selection in the context of neuroimaging analysis, a novel contribution implemented through {\tt fastHDMI}. Such advancements are critical for dissecting the complex architectures inherent in neuroimaging datasets, offering refined mechanisms for variable selection against the backdrop of high dimensionality. Employing the preprocessed Autism Brain Imaging Data Exchange (ABIDE) dataset \cite{Cameron2013, Barry2020} as a foundation, we assess the efficacy of these variable screening methodologies through extensive simulation studies. These evaluations encompass a diverse set of conditions, including linear and nonlinear associations, alongside continuous and binary outcomes. The results delineate the {\em Fast Fourier Transform Kernel Density Estimation (FFTKDE)}-based mutual information estimation approach as preeminent for feature screening with continuous nonlinear outcomes, while the binning-based methodology is identified as superior for binary outcomes contingent on nonlinear underlying probability preimage. For linear simulations, a parity in performance is observed for continuous outcomes between the absolute Pearson correlation and FFTKDE-based mutual information estimation, with the former also exhibiting dominance in binary outcomes predicated on linear underlying probability preimage. A comprehensive case analysis utilizing the preprocessed Autism Brain Imaging Data Exchange (ABIDE) dataset further illuminates the applicative potential of {\tt fastHDMI}, demonstrating the predictive capabilities of models constructed from variables selected through our implemented screening methods. This research not only substantiates the computational prowess and methodological robustness of {\tt fastHDMI}, but also contributes significantly to the arsenal of analytical tools available for neuroimaging research. }




\maketitle

\section{Introduction }\label{sec:introduction}

The question of how to best select a subset of variables from a large set is a commonly investigated topic in high--dimensional model fitting \cite{Chandrashekar2014}. This topic is often called ``variable selection'' in statistics, or ``feature selection'' in the machine learning world. Feature selection may be necessary either to fit a particular statistical model or, in some situations, because the data are too large for memory. Neuroimaging data provide a good example of such challenges. For example, \ac{MRI} result in measurements at millions of voxels \cite{Bell2018, Liang2022, Linn2016, Fan2016}, and the development of multiple imaging modalities is leading to multiple high--dimensional sets of features, each capturing a different aspect of brain function, that can show widespread correlation patterns within and between each modality.
These high dimensions in neuroimaging data have stimulated the development of variable selection methods; indeed, there has been a recent surge in publications: see, for example,
\cite{Adeli2017,Fan2016,Febles2022,GomezVerdejo2019,Hao2020,He2018,Hunt2014,Ivanoska2021,Mohr2006,Martino2008,Pereda2018,Roy2021,Schloegl2002,Sofer2014,Suresh2022}. These papers take a wide variety of strategies ranging from univariate to multivariate selection. Among these, \citeauthor{Fan2016, Schloegl2002}~\cite{Fan2016, Schloegl2002} considered absolute correlation or mutual information with respect to the outcome as a conventional univariate approach; selection based on sparse-inducing penalties on multi-variable models were proposed on the data \cite{Fan2016, Hao2020, Hunt2014, Roy2021} or transformed data \cite{Adeli2017}. Multivariate selection based on random forest variable importance \cite{Febles2022, Hao2020} or sign consistency from the support vector machine \cite{GomezVerdejo2019} has also been applied previously.  A ``potential support vector machine" was applied by \citeauthor{Mohr2006}~\cite{Mohr2006}, an idea that rests on exchanging   the roles of data points and features. Another approach can be seen in \cite{Martino2008}, where they selected features recursively based on multivariate model fitting. Evidently, these papers take a wide variety of strategies ranging from simple methods like analyzing the direct absolute correlation between outcomes and features, to more complex approaches involving the use of univariate regression coefficients, univariate copulas, and techniques that leverage variable importance measures or sparse penalties in multivariate model fitting. Variable selection under a multivariable model generally requires certain assumptions, often including the assumption of linearity, which is not robust to misspecification. Furthermore, variable selection based on marginal associations demands less computational power and memory and can easily adapt to data inflow. Additionally, variable selection within a joint model framework allows for variable screening conditioned on other covariates, such as confounders. 

Although Pearson correlation is frequently used to measure the association between covariates and the outcome, in situations where nonlinearity may be present, a variety of strategies have been introduced to examine the relationship between the outcome and the covariates \cite{Renyi1959, Reshef2011, Speed2011}. These methods, when utilized for feature screening, effectively equate to screening via mutual information, as they are all deterministic monotonically increasing functions of mutual information. Among the strategies for feature selection, an entropy-based method, \emph{mutual information} has two appealing characteristics. As defined in \eqref{eq:mi-defn}, mutual information is defined as the \ac{KL divergence} between the joint distribution of two variables and their outer product distribution, effectively quantifying their dependency. This method can carry out model-independent feature selection, and is robust to non-linearity between the outcome and the features. For these reasons, mutual information has already been a popular choice for neuroimaging data. \citeauthor{Ince2016}~\cite{Ince2016} proposed to estimate mutual information based on the Gaussian Copula for continuous data, which works well for approximately Gaussian data, such as local field potentials and M/EEG data \cite{Magri2009}. \citeauthor{Nemirovsky2023}~\cite{Nemirovsky2023} advanced the analysis of functional MRI data by implementing integrated information theory, which is calculated based on the mutual information between the state of the conscious system over time and across the conscious system's partitions. 
\citeauthor{Tsai1999}~\cite{Tsai1999} used mutual information to analyze functional MRI data to compute an activation map. \citeauthor{Schloegl2002}~\cite{Schloegl2002} used mutual information to study the EEG-based brain-computer interface. \citeauthor{Chai2009}~\cite{Chai2009} and \citeauthor{Li2022}~\cite{Li2022} employed multivariate mutual information to study functional connectivity between brain regions in functional MRI data. \citeauthor{Combrisson2022}~\cite{Combrisson2022} proposed a nonparametric permutation-based framework for neurophysiological data to analyze cognitive brain networks.  

While mutual information estimation for discrete random variables is trivial, the estimation of mutual information for continuous random variables can be done using a few different approaches. 
One fundamental method is to estimate mutual information based on the binning of continuous variables to treat them as discrete variables. \citeauthor{Steuer2002}~\cite{Steuer2002} reported improved performance using \ac{KDE} based methods. \ac{KDE}-based methods numerically calculate the mutual information estimation based on the estimated kernel density functions \cite{Moon1995}.  The \ac{kNN} approach was previously adapted to estimate mutual information \cite{Faivishevsky2008,Kraskov2004,Victor2002,Pal2010,Lord2018,Gao2014}. \citeauthor{Khan2007}~\cite{Khan2007} compared the performance of mutual information estimators based on \ac{kNN} and \ac{KDE} and concluded that \ac{KDE}-based mutual information estimators outperform \ac{kNN}--based estimators for small samples with a high noise level. \citeauthor{Gao2014}~\cite{Gao2014} argued that accurate estimation of mutual information of two strongly dependent variables using \ac{kNN}-based methods requires a prohibitively large sample size. As shown later in our simulation studies in Section \ref{subsec:ABIDE-simulation}, our \ac{KDE}-based mutual information screening method also outperforms the \ac{kNN}-based counterpart. Since kernel density estimation on large volume of data is a computationally challenging approach and that neuroimaging data is usually of large volume, variable screening based on mutual information has never been implemented for neuroimaging data to the best of our knowledge. In this paper, we implement variable screening methods using a few different approaches and carried out comprehensive simulation and real case studies using the preprocessed ABIDE data \cite{Cameron2013, Barry2020}. The variable screening functionality is encapsulated within our Python package, {\tt fastHDMI}, an acronym for \emph{Fast high--dimensional Mutual Information estimation}. This package is specifically designed to facilitate the effective processing and analysis of substantial volumes of neuroimaging data using a few different computationally efficient estimation methods. 


In Section \ref{sec:mi-estimation}, we will explore the concept of mutual information and provide an overview of the estimation methods. Subsequently, Section \ref{sec:sim-and-case-studies} assesses the efficacy of variable selection and the computational speed of the variable selection methods implemented in our {\tt fastHDMI} package. These methods encompass \emph{Fast Fourier Transform-based Kernel Density Estimation (FFTKDE)} mutual information estimation, mutual information estimation based on binning of continuous variables with the number of bins determined using the results of a previous study \cite{Birge2006} utilizing bounds on the risk of penalized maximum likelihood estimators due to Castellan \cite{Castellan2000}, \ac{kNN}-based mutual information estimation, and Pearson correlation. The \ac{kNN}-based mutual information estimation utilized in our work is adapted from the {\tt scikit-learn} library. We will begin by examining these variable screening methods within our {\tt fastHDMI} package through simulations in Section \ref{subsec:ABIDE-simulation}, then proceed to compare their computing speeds. Finally, in Section \ref{subsec:ABIDE-case-studies}, the performance of the predictive models created with the variables selected using our four implemented methods will be demonstrated. 

\section{Estimation of Mutual Information }\label{sec:mi-estimation}

The entropy-based screening methods are based on Shannon's
entropy \cite{Shannon1948}. Let $\mathbf{X}\in\mathbb{R}^{n}$ denote
a random variable residing in a probability space with probability
mass or density function $p\left(\mathbf{X}\right)$. Shannon's entropy
is defined as 
\begin{equation}
    \label{eq:shannon-entropy}
    H\left(\mathbf{X}\right)\coloneqq\mathbb{E}\left[-\log p\left(\mathbf{X}\right)\right].
\end{equation}
Furthermore, Lebesgue's decomposition theorem expands the above definition
for all other random variables. Relative entropy, also known as the {\em\ac{KL divergence}}, is a specific case of Bregman divergence applied to $-H$, the negative of Shannon’s entropy, which is a strictly convex functional: 
\begin{equation}
D_{KL}\left(\mathbf{X}_{1}\parallel\mathbf{X}_{2}\right)\coloneqq\mathbb{E}_{\mathbf{X}_{1}}\left[-\log\frac{p\left(\mathbf{X}_{2}\right)}{p\left(\mathbf{X}_{1}\right)}\right].
\end{equation}

Moreover, mutual information is defined as the \ac{KL divergence} from
the joint distribution $\left(\mathbf{X},\mathbf{Y}\right)$ to the
outer product distribution $\mathbf{X}\otimes\mathbf{Y}$, hence symmetric.
For random variables $\mathbf{X},\mathbf{Y}$, the mutual information
\begin{equation}
I\left(\mathbf{X},\mathbf{Y}\right)\coloneqq D_{KL}\left(\left(\mathbf{X},\mathbf{Y}\right)\parallel\mathbf{X}\otimes\mathbf{Y}\right).\label{eq:mi-defn}
\end{equation}
$\mathbf{X}$ and $\mathbf{Y}$ in \eqref{eq:mi-defn} are typically
univariate for variable screenings. The implementation of \ac{KDE}--based mutual information
estimation uses \ac{FFT} based \ac{KDE} methods from
the Python package {\tt KDEpy} \cite{Odland2018}. FFT-based \ac{KDE} was initially
proposed by \citeauthor{Silverman1982}~\cite{Silverman1982} on Gaussian kernels with much faster
computing speed and much lower numerical errors.
As shown in the paper, such an approach significantly solves the computational
speed challenges that \ac{KDE} usually faces \cite{Silverman1982}. The performance of \ac{KDE} usually
depends on the bandwidth and kernel selection. While we leave it for
users to choose kernel and bandwidth, the default arguments are
set to be the state-of-the-art \emph{Improved Sheather-Jones }bandwidth
\cite{Botev2010} with Epanechnikov kernel \cite{Epanechnikov1969}. For a detailed explanation of the FFTKDE method for mutual information estimation, see Appendix \ref{apdx:method-contribution}.

At the same time, mutual information estimation using the \ac{kNN} method leverages the \ac{kNN} algorithm for entropy estimation, a technique introduced by \citeauthor{L.F.Kozachenko1987}~\cite{L.F.Kozachenko1987}. This method estimates Shannon entropy, as detailed in equation \eqref{eq:shannon-entropy}, with the sample mean, alongside a trinomial distribution to estimate $\widehat{p\left(x_j\right)}$. The binning approach for mutual information estimation converts continuous variables into discrete variables through binning, with the optimal number of bins guided by findings from a previous study by \citeauthor{Birge2006}~\cite{Birge2006}, which derived the optimal number of bins based on the bounds on the risk of penalized maximum likelihood estimators due to \citeauthor{Castellan2000}~\cite{Castellan2000}. Pearson correlation is calculated through the standardized inner product of outcomes and variables. Additionally, to drastically improve the processing speed for large-scale datasets, our package incorporates multiprocessing capabilities, enabling parallel processing across all employed methods. This adaptation to parallel computing significantly enhances the utility of our package, especially for extensive neuroimaging data analyses.

Previous studies demonstrated that the three density estimation methods discussed in this paper, \ac{KDE}, \ac{kNN}, and histogram--based methods, are consistent estimators under suitable conditions. The Lebesgue integral, as a linear operator, has its boundedness equivalent to continuity in a normed linear space. Since expectation is a linear operator, it is continuous under appropriate norms when it is bounded. By the continuous mapping Theorem, the mutual information estimated using these three density estimators is consistent, as the mutual information functional is continuous with respect to the joint likelihood, and continuity is preserved under finite composition. 

Furthermore, since mutual information is continuous with respect to the joint density, sufficiently small numerical errors will not significantly perturb the mutual information estimation. The numerical error associated with the \ac{FFT} procedure arises from multiple sources beyond numerical precision, including errors from using a finite number of \ac{DFT} terms --- such as discretization, truncation of frequencies, and aliasing; and errors from applying \ac{FFT} to a non--periodic function, including boundary effects, zero--padding, and interpolation. Notably, Fourier’s theorem implies that the error from \ac{FFT} for {\em periodic} functions vanishes asymptotically with respect to the number of \ac{DFT} terms. With a computational complexity of $O\left( n\log n \right)$, utilizing a sufficiently fine grid can mitigate these errors while maintaining high computational efficiency. Moreover, \ac{KDE} is inherently non-periodic. Consequently, errors due to boundary effects, zero–padding, and interpolation are influenced by the chosen interval for \ac{KDE} and will not asymptotically vanish with respect to the number of \ac{DFT} terms. The error due to the chosen bounded interval in which the data points reside presents a general challenge when evaluating mutual information numerically, not limited to the \ac{FFT} approach. Additionally, it is important to note that numerical errors, though generally insignificant when using a large number of \ac{DFT} terms, will not vanish asymptotically with respect to the number of data points in the dataset. In summary, \ac{FFT} is an efficient tool to perform \ac{KDE} while maintaining high computational efficiency, as evidenced by previous studies \cite{Silverman1982}.

\section{Simulation and Case Studies }\label{sec:sim-and-case-studies}
\emph{Autism Brain Imaging Data Exchange
(ABIDE) preprocessed Data} consists of preprocessed functional MRI
brain imaging data from \emph{$539$} individuals suffering from ASD
and $573$ typical controls \cite{Cameron2013}. In this paper, we used the preprocessed ABIDE data consisting of $149955$ brain imaging variables, together with age, biological sex, and diagnosis of autism for $508$ cases and $542$ controls \cite{Cameron2013, Barry2020}. The preprocessing was carried out exactly the same manner as the preprocessing performed earlier by \citeauthor{Barry2020}~\cite{Barry2020} (see also \cite{Fischl2012, Dale1999}): the T1-weighted Magnetic Resonance scans were processed through the FreeSurfer 6.0 pipeline \cite{Fischl2012} on the CBrain computing facility \cite{Sherif2014}. This pipeline delineates the cortical surface from magnetic resonance scans, allowing the quantification of the cortical thickness across the brain hemispheres \cite{Fischl2012, Dale1999}. The process involves several steps: affine registration to MNI305 space \cite{Collins1994}, bias field correction, removal of non-cortical regions, and the estimation of white matter and pial surfaces from intensity gradients, which are used to estimate cortical thickness. These cortical surfaces are projected into a common space (fsaverage) for comparison across individuals.

Brain MRI data has been used to predict age to study the brain aging process linked to diseases such as Alzheimer’s disease and Parkinson’s disease \cite{Jonsson2019,Jiang2020,Cole2017,Franke2010,Liem2017}. For the case studies based on the preprocessed ABIDE data \cite{Cameron2013, Barry2020} in Section \ref{subsec:ABIDE-case-studies}, we choose
age at the MRI scan as the continuous outcome and autism diagnosis
as the binary outcome. When using age at the MRI scan as the outcome,
we adjust for sex and autism diagnosis; we using autism diagnosis
as the outcome, we adjust for age and sex. 
We compare the few screening methods in our Python package {\tt fastHDMI}, including mutual information estimation using the FFTKDE and \ac{kNN} originally implemented in the {\tt scikit-learn} library, as well as Pearson correlation.


\subsection{Simulation based on the preprocessed ABIDE data \cite{Cameron2013, Barry2020} \label{subsec:ABIDE-simulation} }

We decided to simulate outcomes based on the preprocessed ABIDE MRI features in order to preserve the distribution patterns and the correlation structure in this high--dimensional dataset. Therefore, we simulated both nonlinear and linear outcomes from the preprocessed
ABIDE data \cite{Cameron2013, Barry2020}. Let $\mathbf{X}\in\mathbb{R}^{N\times p}$
denote the design matrix; i.e., all the MRI brain imaging variables from
the entire preprocessed ABIDE dataset. The simulation of the \emph{nonlinear}
outcomes proceeds in this manner -- the nonlinearity for continuous
outcomes comes from the quadratic manipulation, i.e., step 4: 

\vspace{.1in}

\begin{tcolorbox}[breakable]
\vspace{.1in}
\begin{enumerate}
\item Pick the number of ``true'' covariates $p_{\text{true}}$, choose $p_{\text{true}}$ uniformly randomly from the full feature set; let
$\mathbf{X}_{\text{true}}\in\mathbb{R}^{N\times p_{\text{true}}}$ 
denote the corresponding design sub-matrix.
\item Simulate the corresponding ``true'' coefficients $\boldsymbol{\beta}_{\text{true}}\in\mathbb{R}^{p_{\text{true}}}$ with $\boldsymbol{\beta}_{\text{true}}\sim N_{p_{\text{true}}}\left(1,\Sigma_{\boldsymbol{\beta}_{\text{true}}}\right)$ and $\Sigma_{\boldsymbol{\beta}_{\text{true}}}$ being a $0.6$ Toeplitz matrix. The correlation design aims to replicate the phenomenon of correlated brain signals. 
\item Standardize the design sub-matrix for the true features $\mathbf{X}_{\text{true}}$, to obtain
$\mathbf{X}_{\text{true},1}$.
\item For nonlinear simulations only: take the element-wise square of $\mathbf{X}_{\text{true},1}$
and then standardize the matrix again to obtain $\mathbf{X}_{\text{true},2}$; the standardization here is to ensure that each feature impacts the simulated outcome proportionally.
\item The continuous and binary outcomes are then simulated in this manner: 
\begin{enumerate}
\item To simulate continuous outcomes: 
\begin{enumerate}
\item Pick $\text{SNR}=3$; calculate $\sigma_{\text{true}}=\sqrt{\frac{\boldsymbol{\beta}_{\text{true}}^{T}\mathbf{X}_{\text{true},2}^{T}\mathbf{X}_{\text{true},2}\boldsymbol{\beta}_{\text{true}}}{\text{SNR}}}$; 
\item Simulate the error $\varepsilon_{j}\overset{i.i.d.}{\sim}N\left(0,\sigma_{\text{true}}^{2}\right)$;
\item The outcome is simulated as $\mathbf{y}=\mathbf{X}_{\text{true},2}\boldsymbol{\beta}_{\text{true}}+\boldsymbol{\varepsilon}$. 
\end{enumerate}
\item To simulate binary outcomes: 
\begin{enumerate}
\item Calculate $\boldsymbol{\tau}=\mathbf{X}_{\text{true},2}\boldsymbol{\beta}_{\text{true}}$; 
\item Standardize $\boldsymbol{\tau}$, obtain $\boldsymbol{\tau}^{\prime}$
-- this is to avoid the data being too centered, which will cause
all simulated binary outcomes in the same class; 
\item Take $\boldsymbol{\tau}^{\prime\prime}=\boldsymbol{\tau}^{\prime}+\text{arctanh}\sqrt{\frac{1}{3}}$
for \emph{translated} binary outcome simulations, or $\boldsymbol{\tau}^{\prime\prime}=\boldsymbol{\tau}^{\prime}$
for \emph{original} binary outcome simulations. The translated binary outcome
simulation is to make the logistic transformation of centered data in the next step
as nonlinear as possible, as $\text{\ensuremath{\pm}arctanh\ensuremath{\sqrt{\frac{1}{3}}}}$
is the location for the logistic transformation to achieve the greatest
absolute curvature value; 
\item The binary outcome is then simulated as $y_{j}\overset{\text{indep.}}{\sim}\text{Bern}\left(\text{logistic}\left(\tau_{j}^{\prime\prime}\right)\right)$. 
\end{enumerate}
\end{enumerate}
\end{enumerate}
For linear simulations, we omit step 4 and take $\mathbf{X}_{\text{true},2}\coloneqq\mathbf{X}_{\text{true},1}$
thereafter. 
\vspace{.1in}
\end{tcolorbox}

\vspace{.05in}

The screening of features with respect to the simulated continuous and binary outcomes $\mathbf{y}$ are then carried out using the original entire design matrix $\mathbf{X}$. 
Variable selection performance is measured by \emph{Variable Selection
Area under Receiver Operating Curve (AUROC)}, which is the AUROC calculated with the true labels taking value $1$
for the simulated true coefficients and $0$ for other coefficients,
and the ranking of the coefficients follows the absolute value of
the three association measures, respectively; i.e., $\widehat{MI}$
based on FFTKDE and \ac{kNN}, as well as Pearson correlation. The top $p_{\text{true}}$ of the most associated covariates are then taken as selected covariates, which will take value $1$, and the others will take value $0$. Variable Selection
AUROC therefore measures the matching between the selected covariates and the simulated ``true'' covariates. Such
measures can differentiate distinct methods when the traditional measures
such as classification rate or adjusted Rand Index can not -- a scenario frequently occurs to variable selection for ultra-high--dimensional
data. 

\begin{figure}[H]
\centering{}\includegraphics[width=1.0\textwidth]{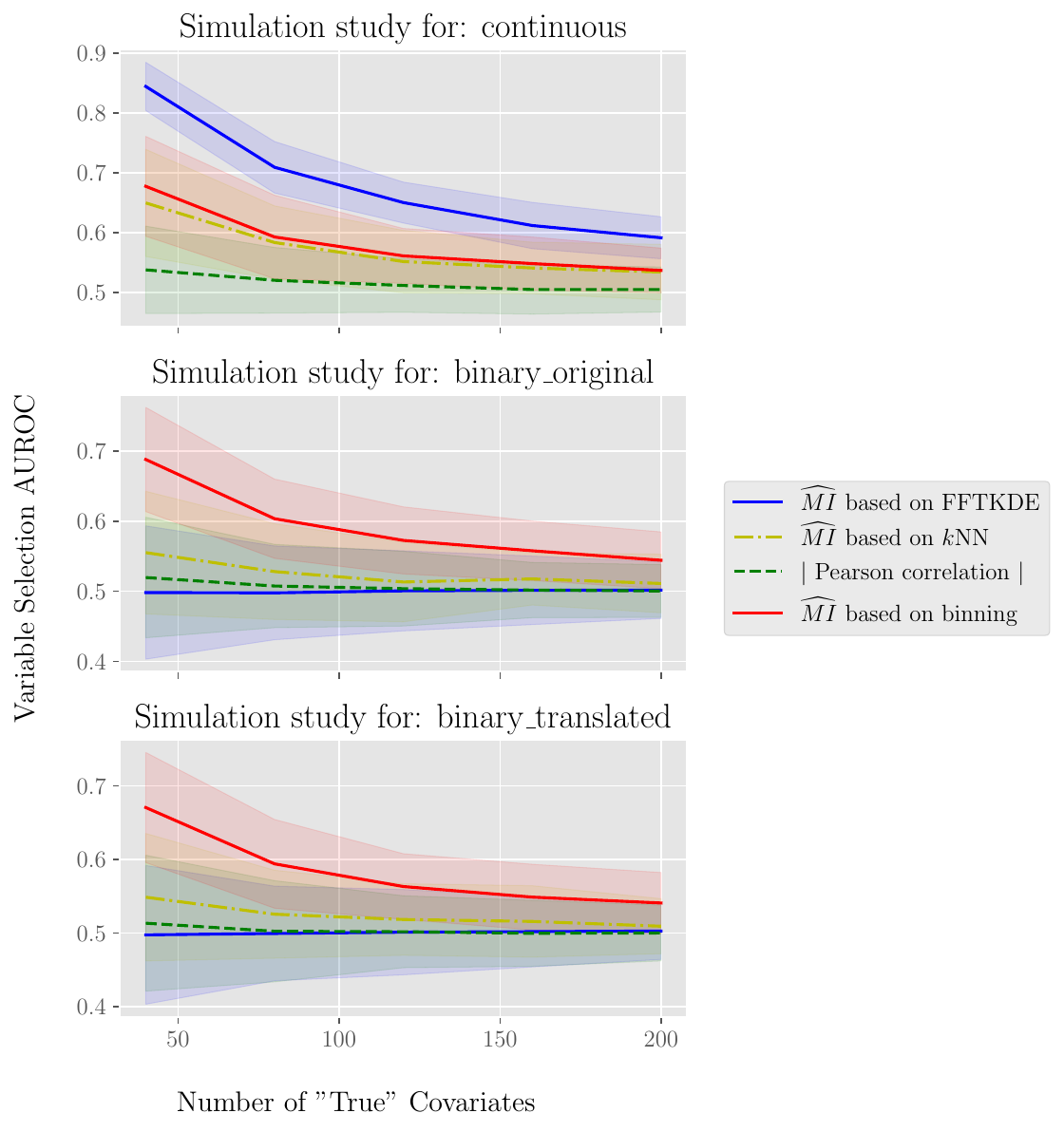}\caption{Variable selection AUROC on the simulated \emph{nonlinear} continuous
and original/translated binary outcomes; the horizontal axis is the
number of \textquotedblleft true\textquotedblright{} covariates used
in the outcome simulation. Means with their $95\%$ confidence intervals were plotted
for $100$ simulation replications. \label{fig:Variable-selection-AUROC-nonlinear}}
\end{figure}

\begin{figure}[H]
\centering{}\includegraphics[width=1.0\textwidth]{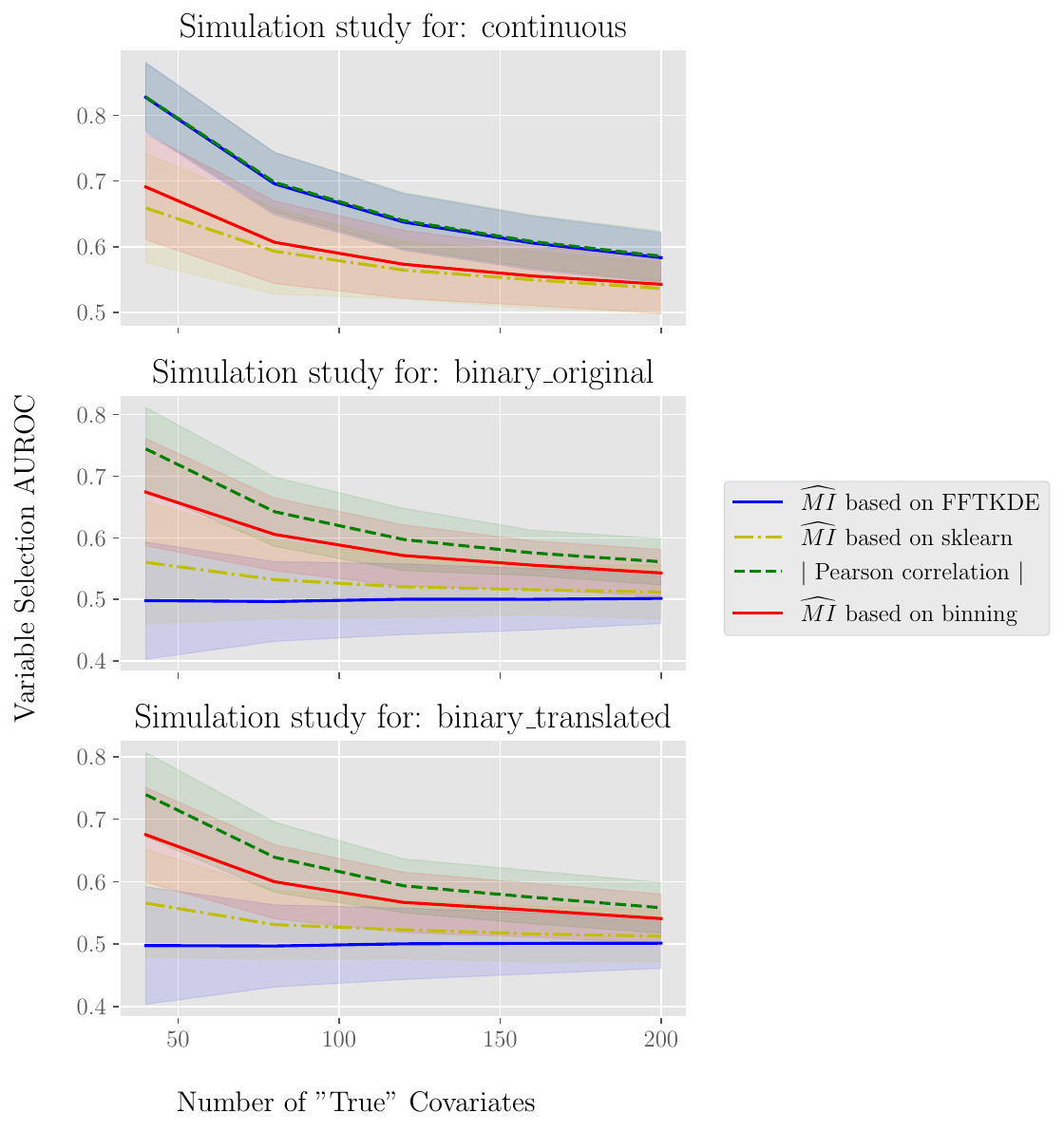}\caption{Variable selection AUROC on the simulated \emph{linear} continuous
and original/translated binary outcomes; the horizontal axis is the
number of \textquotedblleft true\textquotedblright{} covariates used
in the outcome simulation. Means with their $95\%$ confidence intervals were plotted
for $100$ simulation replications. \label{fig:Variable-selection-AUROC-linear}}
\end{figure}

We evaluate the efficacy of our implemented variable screening methods in {\tt fastHDMI} package, including: 1) Mutual information estimation using FFTKDE, 2) Mutual information estimation using \ac{kNN}, 3) Mutual information estimation through binning, and 4) absolute Pearson correlation. Our findings, illustrated in Figures \ref{fig:Variable-selection-AUROC-nonlinear} and \ref{fig:Variable-selection-AUROC-linear}, reveal that for continuous outcomes, the FFTKDE-based mutual information estimator outperforms its counterparts. In scenarios with linear relationships, FFTKDE-based mutual information estimator and absolute Pearson correlation are jointly the most effective. Conversely, for binary outcomes, the binning-based mutual information estimator excels in capturing nonlinear associations, whereas other methodologies display substantially overlapping confidence intervals. In linear association contexts, Pearson correlation emerges as the most effective method for binary outcomes. Interestingly, Pearson correlation, particularly when employed with a balanced number of cases and controls, inherently correlates to a two-sample testing approach, which explains its superior performance for binary outcomes with linearly simulated underlying probability pre-image. 

\begin{figure}[H]
\centering{}\includegraphics[width=1.0\textwidth]{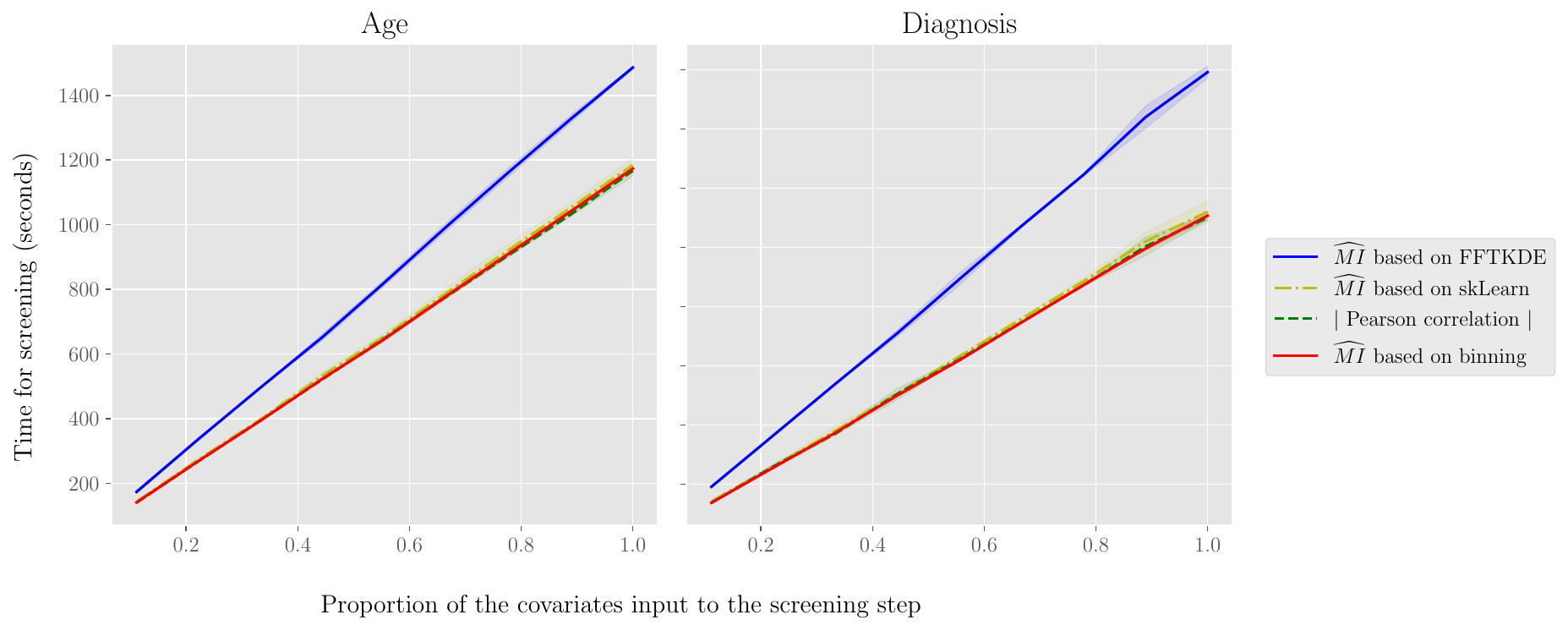}\caption{Running speeds of variable screening for continuous (age) and binary (diagnosis) outcomes utilizing the methods under study. The horizontal axis represents the proportion of features introduced into the screening phase, while the vertical axis measures the time in seconds to complete the screening. The plot displays the mean running times and their corresponding $95\%$ confidence intervals (C.I,), derived from 5 simulation replications. \label{fig:running-speed}}
\end{figure}

All discussed variable screening methods were conducted concurrently on $16$-core CPUs on Compute Canada. The fast Fourier transform (FFT) algorithm is leveraged to significantly enhance the efficiency of the \ac{KDE} estimation process, traditionally viewed as computationally intensive. As depicted in Figure \ref{fig:running-speed}, the execution times to complete the screenings with all the methods implemented in our {\tt fastHDMI} package are assessed. Notably, the \ac{KDE}-based mutual information estimation, often anticipated to be slower, exhibited competitive speed akin to alternative methods, courtesy of the FFT algorithm's effectiveness. This computational efficiency was achieved with the same CPU configuration, while intentionally avoiding multiple data duplications in memory during multiprocessing. Given the substantial size of high--dimensional datasets, duplicating such datasets in memory is generally impractical.

\subsection{Preprocessed ABIDE data case studies \cite{Cameron2013, Barry2020} -- predict age and diagnosis \label{subsec:ABIDE-case-studies} }


In this subsection, we evaluate the performance of various variable screening techniques implemented in the {\tt fastHDMI} package using preprocessed ABIDE data \cite{Cameron2013, Barry2020}. Initially, we deploy the four variable screening methods to identify the features most associated with the outcome. Since we are fitting multiple penalized models, standardization of the selected variables is carried out to achieve a sample mean of $0$ and a standard deviation of $1$. This step is crucial for ensuring consistent penalization across all coefficients of the penalized covariates.

Subsequently, we divide the dataset, stratified by the outcome, into a training set comprising $80\%$ of the observations and a testing set with the remaining $20\%$. This stratification ensures a balanced representation of the outcomes in both sets. For the continuous outcome, age, we employ binning to categorize observations into $30$ bins based on their outcome values, followed by stratification based on the bin labels. This approach allows for the division of the dataset into training and testing sets with similar outcome means, an important factor for reliable prediction performance comparison.

For the continuous outcome variable, age at MRI scan, we fit several models: elastic net, least-angle regression (LARS), least absolute shrinkage and selection operator (LASSO), LASSO-LARS, linear model, Random Forest regressor, and ridge regression. Except for the Random Forest regressor, which utilizes the out-of-bag error scored by $R^2$ for model averaging, all models are tuned using $5$-fold cross-validation with validation set $R^2$ as the scoring function for penalty hyperparameters.

For binary outcomes, diagnosis of autism disorder, we fit both unpenalized and penalized logistic regressions (using $\ell_1$, $\ell_2$, and elastic net penalties), as well as the Random Forest classifier. All models, with the exception of the Random Forest classifier, which uses out-of-bag error scored by Gini impurity for model averaging, are tuned using $5$-fold cross-validation, scored by mean accuracy for the penalty hyperparameters.

Unlike simulation studies in Section \ref{subsec:ABIDE-simulation}, where ``true'' signals are known, case studies lack such definitive benchmarks, necessitating reliance on model-based performance metrics. Hence, we use testing set $R^2$ for continuous outcomes and testing set Area Under the Receiver Operating Characteristic (AUROC) for binary outcomes to evaluate model performance. 

\begin{figure}[H]
\includegraphics[width=0.95\textwidth]{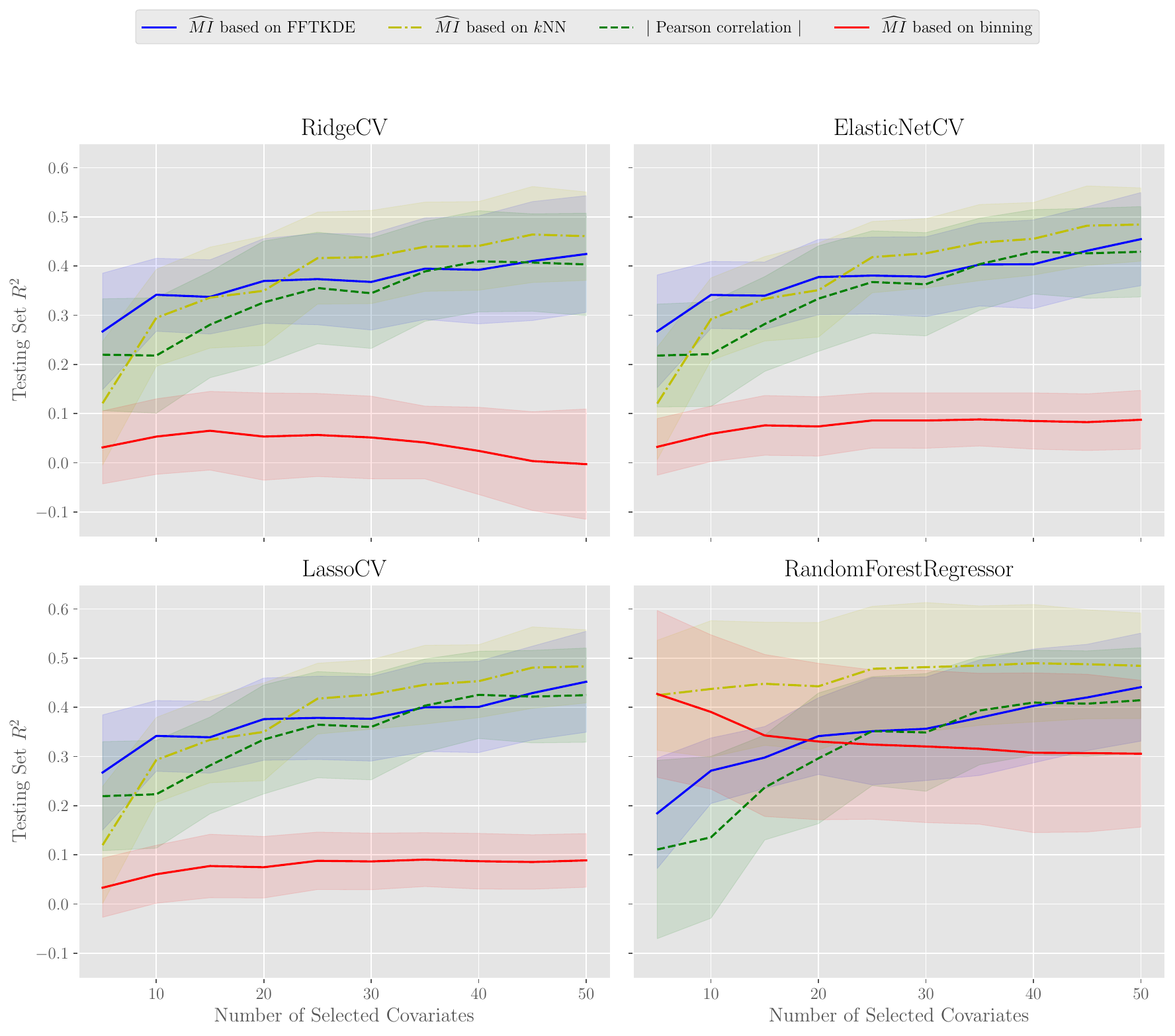}
\caption{Testing Set $R^{2}$ for age at the scan outcome v.s. the number of
most associated brain imaging covariates based on the association
measure rankings. Means with their $95\%$ confidence intervals were plotted for $20$
simulation replications. \label{fig:Case-Study-continuous}}
\end{figure}

\begin{figure}[H]
\centering{}
\includegraphics[width=0.95\textwidth]{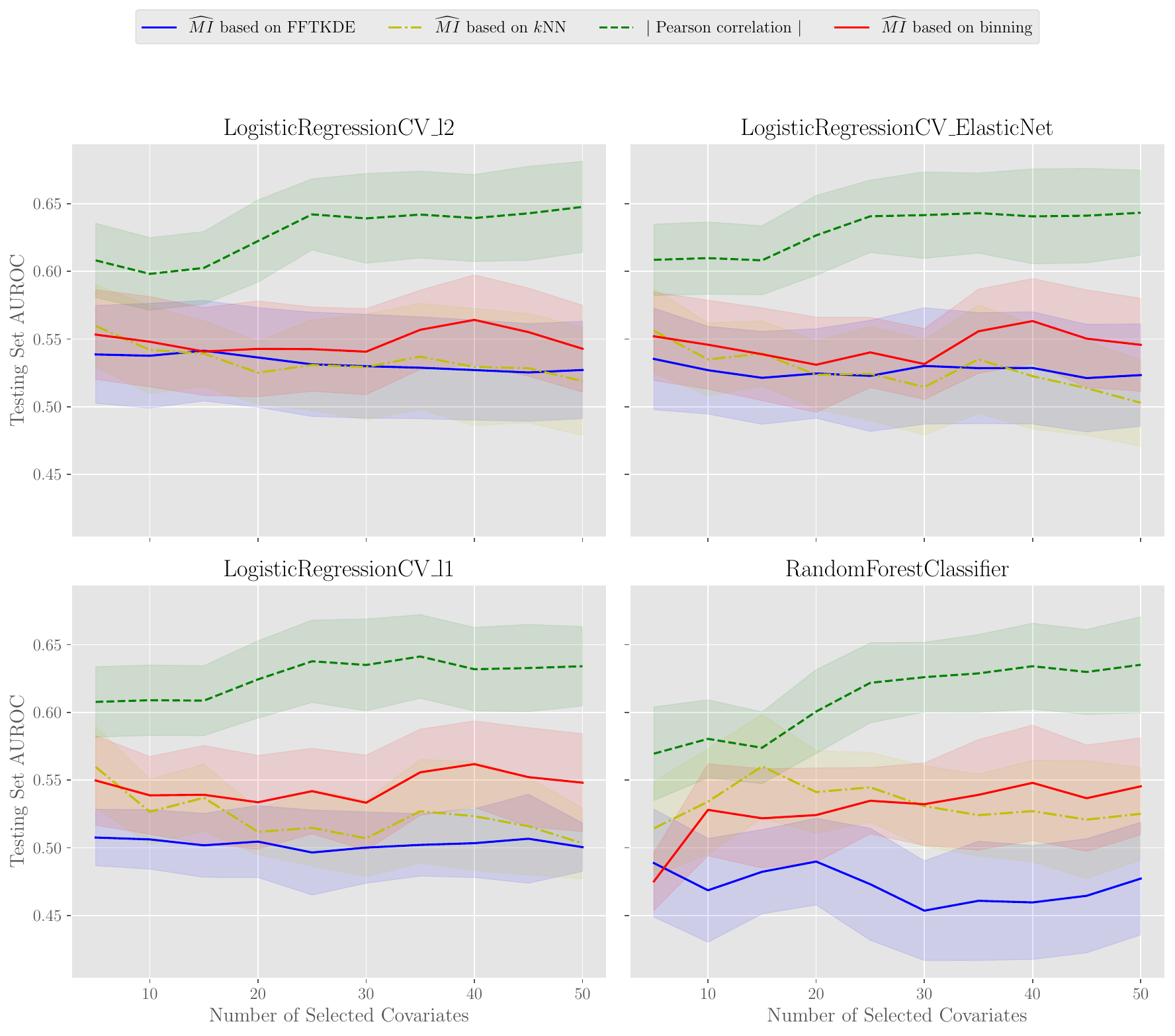}
\caption{Testing Set AUROC for autism diagnosis outcome v.s. the number of
most associated brain imaging covariates based on the association
measure rankings. Means with their $95\%$ confidence intervals were plotted for $20$
simulation replications. \label{fig:Case-Study-binary}}
\end{figure}

Figure \ref{fig:Case-Study-continuous} illustrates that in predicting the continuous outcome, age at MRI scan, linear models utilizing brain imaging variables selected using mutual information estimations via FFTKDE or \ac{kNN} emerge as the best-performing. Conversely, models built using variables selected by mutual information estimations based on binning exhibit the least predictive capability. However, within the context of random forest regression, models built using variables chosen through mutual information estimation by \ac{kNN} outperform the rest. Figure \ref{fig:Case-Study-binary} indicates that for the binary outcome of autism diagnosis, models constructed with variables selected via absolute Pearson correlation yield superior predictive performance. This phenomenon could stem from multiple factors, including the linear nature of the assessment model, which favors linear association measures, or a linear relationship between age at MRI scan, the probability of autism diagnosis, and the brain imaging covariates.

\section{Conclusion and Discussion }

In this paper, we introduce the Python package {\tt fastHDMI}, designed to streamline variable screening through three distinct mutual information estimation methods along with absolute Pearson correlation. Our evaluations, conducted on the large, high--dimensional preprocessed ABIDE data \cite{Cameron2013, Barry2020}, affirm {\tt fastHDMI}'s computational efficiency and robustness. Through extensive simulation studies, which encompass both simulations for linear and nonlinear associations, as well as continuous and binary simulated outcomes, we evaluated the performance of each implemented variable screening method. Our findings reveal that for simulated continuous nonlinear outcomes, the FFTKDE-based mutual information estimation method excels in variable selection. Similarly, for simulated binary outcomes with a nonlinear underlying probability preimage, the binning-based mutual information estimation stands out. In the cases of simulated continuous linear outcomes, both absolute Pearson correlation and FFTKDE-based mutual information estimation share the top performance. Furthermore, absolute Pearson correlation is superior for binary outcomes simulated with linear underlying probability preimage. Complementing our simulations, a comprehensive case study on the preprocessed ABIDE data \cite{Cameron2013, Barry2020} showcased the predictive capabilities of models crafted from the most relevant covariates identified by our methods. By pioneering sophisticated variable selection techniques in the domain of high--dimensional neuroimaging data, our work stands as a critical advancement, fostering novel pathways for research exploration and analytical insight within the scientific community. A promising avenue for future research could be to explore variable screening based on non-parametric copula models \cite{Rabhi2019}.

\section{Disclaimer}
All codes to reproduce the simulation and case study results of this paper and outputs from Calcul Quebec/Compute Canada can be found on the following GitHub repository:

\href{https://github.com/Kaiyangshi-Ito/fastHDMI}{https://github.com/Kaiyangshi-Ito/fastHDMI}

\section{Acknowledgments}
This work was supported by the ISM Scholarship for Outstanding PhD Candidates awarded to K. Yang, the NSERC Discovery Grant to C. Greenwood (Grant Number: RGPIN-2019-04482), the NSERC Discovery Grant to M. Asgharian (Grant Number: RGPIN-2024-05640), and the CANSSI Collaborative Research Team Grant to C. Greenwood and G. Cohen Freue.

\printbibliography

\begin{appendices}

\section{Methodology Consideration \label{apdx:method-contribution}}

For a function $f$ defined over an Euclidean space $\mathbb{R}^{n}$, its (continuous) Fourier transform is defined as 
\begin{equation}    \left(\mathcal{F}f\right)\left(\xi\right)\coloneqq\int_{\mathbb{R}^{n}}f\left(x\right)\exp\left(-2\pi i\cdot\left\langle x,\xi\right\rangle \right)dx,
\end{equation}
a linear operator. The Fourier series is then the synthesis formula. Consider a square-integrable function space $L^{2}\left(\left[-\pi,\pi\right]\right)$,
the fundamental results of Fourier analysis \cite{Stein2003} conclude that $\left\{ \phi_{k}\coloneqq\exp\left(ikx\right)\vert k\in\mathbb{Z}\right\} $
is an orthonormal and complete basis for this Hilbert space with the
inner product being defined by 
\begin{equation}
\forall f,g\in L^{2}\left(\left[-\pi,\pi\right]\right),\ \left\langle f,g\right\rangle \coloneqq\frac{1}{2\pi}\int_{-\pi}^{\pi}f\left(x\right)\bar{g}\left(x\right)dx.
\notag
\end{equation}
We remark the that inner product for a complex Hilbert space is linear for the first argument and anti-linear for the second argument. The Fourier series that represents any function $f\in L^{2}\left(\left[-\pi,\pi\right]\right)$
is then 
\begin{equation}
f=\sum_{k=-\infty}^{\infty}\left\langle f,\phi_{k}\right\rangle \phi_{k}.
\notag
\end{equation}
Clearly, ($1D$ continuous) Fourier transform is to extend the idea of decomposing functions on the interval $\left[-\pi,\pi\right]$ to analyzing them across $\mathbb{R}$ by scaling the frequency domain. This approach applies analogously to higher-dimensional situations. The completeness of the Fourier basis is given by the Fourier theorem, while the uniqueness of continuous Fourier transform and the inverse Fourier transform under certain conditions is a key result in Fourier analysis \cite{Stein2003}. An important property of the Fourier series/continuous Fourier transform is the convolution property:
\begin{equation}
\forall f,g\in L^{2}\left(\left[-\pi,\pi\right]\right),\ \mathcal{F}\left(f*g\right)=\left(\mathcal{F}f\right)\cdot\left(\mathcal{F}g\right),
\notag
\end{equation}
where $\mathcal{F}$ denotes the Fourier transform. 

For a finite number of data points, \emph{discrete Fourier transform
(\ac{DFT})} can be used to approximate a function using the Fourier basis
$\left\{ \phi_{k}\right\} $ mentioned above. In the context of our discussion of \ac{DFT}, for a slight abuse of notions, let $\mathcal{F}$ also represent the Fourier series. In physical space, the equispaced grid of points is usually scaled first to match the domain of the \ac{DFT} transform, often chosen as $\left[-\pi,\pi\right]$ for $1D$ data or $\left[-\pi,\pi\right]\times\left[-\pi,\pi\right]$ for $2D$ data. \ac{DFT} then transforms
the function values evaluated at the equispaced data points in the
physical space to Fourier coefficients in the frequency space by multiplication
of the following matrix, called \ac{DFT} matrix: 
\begin{equation}
\Psi\coloneqq N^{-\frac{1}{2}}\left[\begin{array}{ccccc}
\psi^{0} & \psi^{0} & \psi^{0} & \cdots & \psi^{0}\\
\psi^{0} & \psi & \psi^{2} &  & \psi^{N-1}\\
\psi^{0} & \psi^{2} & \psi^{4} &  & \psi^{2\left(N-1\right)}\\
 & \vdots &  & \ddots & \vdots\\
\psi^{0} & \psi^{N-1} & \psi^{2\left(N-1\right)} & \cdots & \psi^{\left(N-1\right)\left(N-1\right)}
\end{array}\right],
\notag
\end{equation}
where $\psi\coloneqq\exp\left(-\frac{1}{N}2\pi i\right)$. \ac{FFT} is an algorithm to efficiently perform the \ac{DFT} for
a finite number of data points, reducing the complexity from $O\left(N^{2}\right)$
to $O\left(N\log N\right)$ \cite{Cooley1965}. Inverse \ac{FFT} can
be done similarly. 

In a two-dimensional space, the \ac{DFT} of the function $f$ is based on the
projection on a $2D$ Fourier basis $\left\{ \phi_{k}\coloneqq\exp\left(ikx+ijy\right)\vert k,j\in\mathbb{Z}\right\} $.
The convolution property and \ac{FFT} in a $2D$ space is then similar
to that of the $1D$ space \cite{Stein2003, Cooley1965}. 

Based on above, kernel density estimation can be computed efficiently
using the convolution property of Fourier transform and \ac{FFT} \cite{Silverman1982}. \citeauthor{Silverman1982}~\cite{Silverman1982} further demonstrated the outstanding numerical performance of \ac{FFTKDE}.
Specifically, the kernel density estimation for $N$ data points is
\begin{equation}
\hat{f}\left(x;\Omega\right)\coloneqq N^{-1}\sum_{j=1}^{N}K\left(x-x_{j};\Omega\right),
\notag
\end{equation}
where $K$ denotes the kernel and $\Omega$ denotes the bandwidth
matrix. Thus, KDE can be carried out efficiently by
\begin{equation}
\hat{f}\left(x;\Omega\right)=N^{-1}\sum_{j=1}^{N}K\left(x;\Omega\right)*\delta\left(x-x_{j}\right),
\notag
\end{equation}
where $\delta$ is Dirac delta, which functions as a ``spike'' and
has Fourier transform being a constant function depending only on the chosen normalization constant of the Fourier transform. This allows $\hat{f}$
to be calculated efficiently, since the convolution property of Fourier transform implies
that 
\begin{equation}
\mathcal{F}\left(\hat{f}\right)\left(x;\Omega\right)=\mathcal{F}\left(K\right)\left(x;\Omega\right)\cdot\mathcal{F}\left(\delta\right)\left(x-x_{j}\right).
\notag
\end{equation}
Then, $\hat{f}\left(x;\Omega\right)$ evaluated on a $2D$ equispaced grid
can be calculated using I\ac{FFT}. Therefore, the evaluated density value on the $2D$ equispaced grid can
be used to calculate the mutual information estimation, specifically, 
\begin{equation}
\widehat{MI}\left(Y,X_{j}\right)=\int_{\text{supp}\left(Y\right)}\int_{\text{supp}\left(X_{j}\right)}\hat{f}_{Y,X_{j}}\left(y,x_{j}\right)\cdot\log\frac{\hat{f}_{Y,X_{j}}\left(y,x_{j}\right)}{\hat{f}_{Y}\left(y\right)\cdot\hat{f}_{X_{j}}\left(x_{j}\right)}dx_{j}dy
\label{eq:numerical-FFTKDE-MI-estimator}
\end{equation}

In \eqref{eq:numerical-FFTKDE-MI-estimator}, $\hat{f}_{Y}\left(y\right)$, $\hat{f}_{X_{j}}\left(x_{j}\right)$, and the expectation estimator itself can be numerically computed using the forward Euler method. Notably, employing the \ac{FFT} for the integration of density functions often fails to deliver satisfactory numerical results, primarily attributed to the inherent periodic characteristics of the method. \eqref{eq:numerical-FFTKDE-MI-estimator} is the equation that we use to calculate the \ac{FFTKDE} mutual information estimator. 

The estimation of mutual information using another nonparametric method, $k$NN \cite{Faivishevsky2008,Kraskov2004,Victor2002,Pal2010,Lord2018,Gao2014}, was also discussed in the paper.
The estimation of mutual information based on $k$NN can be viewed through
the lens of $k$NN density estimator. The bivariate $k$NN density
estimator can be given by 
\[
\hat{f}\left(x;k\right)\coloneqq\frac{k}{N}\cdot\left(\pi\cdot R^{2}\left(x;k\right)\right)^{-1},
\]
where $R\left(x;k\right)$ denotes the Euclidean distance from $x$
to its $k$-nearest-neighbor. In the context of a bivariate density estimator,
$\pi\cdot R^{2}\left(x;k\right)$ represents the area of the Euclidean-normed
closed ball centered at $x$ that includes the $k$-nearest-neighbors
of $x$. Following the idea of empirical CDF, the probability that
a data point is included in this closed ball is $\frac{k}{N}$;
assuming that the density inside the closed ball remains constant, the estimate of such density will be the probability of being included
in the closed ball divided by the area of the closed ball, which is
the bivariate density estimator described above. The multivariate case
with more than two variables can be established in a similar way. 

\end{appendices}



\end{document}